\documentclass[sigconf]{acmart}
\AtBeginDocument{%
  }

\usepackage{marvosym}
\usepackage[most]{tcolorbox}
\usepackage{pifont}

\usepackage{newunicodechar}
\newunicodechar{★}{\ding{72}}

\setcopyright{acmlicensed}
\copyrightyear{2026}
\acmYear{2026}
\setcopyright{cc}
\setcctype{by-nc-nd}
\acmConference[SIGIR '26]{Proceedings of the 49th International ACM SIGIR Conference on Research and Development in Information Retrieval}{July 20--24, 2026}{Melbourne, VIC, Australia}
\acmBooktitle{Proceedings of the 49th International ACM SIGIR Conference on Research and Development in Information Retrieval (SIGIR '26), July 20--24, 2026, Melbourne, VIC, Australia}
\acmDOI{10.1145/3805712.3808597}
\acmISBN{979-8-4007-2599-9/2026/07}

\usepackage{enumitem}

\begin{document}

\title{DeepResearch-9K: A Challenging Benchmark Dataset of Deep-Research Agent}

\author{Tongzhou Wu}
% \orcid{0009-0000-7544-7729}
\affiliation{%
  \institution{Shenzhen Technology University}
  \institution{City University of Hong Kong}
  \city{Shenzhen}
  \country{China}}
\email{tongzhowu3-c@my.cityu.edu.hk}

\author{Yuhao Wang}
\affiliation{%
  \institution{City University of Hong Kong}
  \city{Hong Kong}
  \country{China}}
\email{yhwang25-c@my.cityu.edu.hk}

\author{Xinyu Ma}
\affiliation{%
  \institution{Baidu Inc.}
  \city{Beijing}
  \country{China}}
\email{xinyuma2016@gmail.com}

\author{Xiuqiang He}
\affiliation{%
  \institution{Shenzhen Technology University}
  \city{Shenzhen}
  \country{China}}
\email{hexiuqiang@sztu.edu.cn}

\author{Shuaiqiang Wang}
\affiliation{%
  \institution{Baidu Inc.}
  \city{Beijing}
  \country{China}}
\email{shqiang.wang@gmail.com}

\author{Dawei Yin}
\affiliation{%
  \institution{Baidu Inc.}
  \city{Beijing}
  \country{China}}
\email{yindawei@acm.org}

\author{Xiangyu Zhao\texorpdfstring{\Letter}{}}
\thanks{\Letter \text{Corresponding author}}
\affiliation{%
 \institution{City University of Hong Kong}
 \city{Hong Kong}
 \country{China}}
\email{xianzhao@cityu.edu.hk}

\renewcommand{\shortauthors}{Tongzhou Wu et al.}

\begin{abstract}
  Deep-research agents are capable of executing multi-step web exploration, targeted retrieval, and sophisticated question answering. 
  Despite their powerful capabilities, deep-research agents face two critical bottlenecks: (1) the lack of large-scale, challenging datasets with real-world difficulty, and (2) the absence of accessible, open-source frameworks for data synthesis and agent training. To bridge these gaps, we first construct \textbf{DeepResearch-9K}, a large-scale challenging dataset specifically designed for deep-research scenarios built from open-source multi-hop question-answering (QA) datasets via a low-cost autonomous pipeline. Notably, it consists of (1) 9000 questions spanning three difficulty levels from $\mathcal{L}_1$ to $\mathcal{L}_3$ (2) high-quality search trajectories with reasoning chains from Tongyi-DeepResearch-30B-A3B, a state-of-the-art deep-research agent, and (3) verifiable answers. Furthermore, we develop an open-source training framework \textbf{DeepResearch-R1} that supports (1) multi-turn web interactions, (2) different reinforcement learning (RL) approaches, and (3) different reward models such as rule-based outcome reward and LLM-as-Judge feedback. Finally, empirical results demonstrate that agents trained on DeepResearch-9K under our DeepResearch-R1 achieve state-of-the-art results on challenging deep-research benchmarks. We release the DeepResearch-9K dataset on \url{https://huggingface.co/datasets/artillerywu/DeepResearch-9K} and the code of DeepResearch-R1 on \url{https://github.com/Applied-Machine-Learning-Lab/SIGIR2026_DeepResearch-R1}.
\end{abstract}

\begin{CCSXML}
<ccs2012>
   <concept>
       <concept_id>10002951.10003317</concept_id>
       <concept_desc>Information systems~Information retrieval</concept_desc>
       <concept_significance>500</concept_significance>
       </concept>
 </ccs2012>
\end{CCSXML}
\ccsdesc[500]{Information systems~Information retrieval}

\keywords{Deep Research, Agent, Benchmark, Dataset}

\maketitle
\section{Introduction}
\label{sec:introduction}
The rapid evolution of Large Language Models (LLMs)—marked by sharper reasoning and vast context windows—has shifted the focus of agent research toward more autonomous systems~\cite{liu2025large, singh2025openai,wang2024survey, li2025towards}. We are seeing agents tackle increasingly complex workflows, from synthesizing long-form documents to executing multi-step plans~\cite{qian2025scent,wang2024q}. This evolution highlights the demand for robust deep-research capabilities. It is no longer sufficient to retrieve information; agents must now navigate abstract queries by breaking them into logical subproblems, managing search tools, and constantly refining their reasoning paths based on intermediate findings~\cite{zhao2024retrieval,qian2025hawkbench,arslan2024survey,li2025agent4ranking}. The rise of specialized models such as Tongyi‑DeepResearch~\cite{tongyidr}, Websailor~\cite{li2025websailor}, and leading commercial platforms (e.g., Gemini~\cite{comanici2025gemini}, ChatGPT~\cite{singh2023chat}) reflects how quickly these agentic capabilities are maturing.

Despite this progress, the field is bottlenecked by a shortage of challenging deep-research datasets. Early open‑source benchmarks like NQ~\cite{kwiatkowski2019natural}, HotpotQA~\cite{yang2018hotpotqa}, and MuSiQue~\cite{trivedi2022musique} evaluate models' ability to integrate information, which failed to capture the autonomy needed for actual research. However, even with the introduction of web simulations like Search-R1~\cite{jin2025search}, the exploration remains largely constrained. While recent benchmarks like GAIA~\cite{mialon2023gaia} and BrowseComp\cite{chen2025browsecomp} have successfully raised the bar by focusing on tool-calling and expert domains, there remains a distinct lack of environments that force models to handle the multi-step complexity of a deep-research trajectory~\cite{xu2025harnessing, gao2025llm4rerank}.

Moreover, existing datasets that aim to support deep-research suffer from the following limitations:
\begin{itemize}[leftmargin=*]
    \item \textbf{Insufficient Reasoning Chains:} Most existing datasets require only a few search counts, failing to mirror the extended logical chains characteristic of real-world research~\cite{ma2024ex}. As a result, current models operate within their reasoning capacity, with little incentive to extend beyond shallow, few-step inference.~\cite{li2025survey, zhang2026evoking}.
    \item \textbf{The Search Tool Call Frequency Gap:} While benchmarks like HotpotQA effectively evaluate linguistic diversity and logical composition, they lack systematic difficulty scaling tied to search count~\cite{zhang2026search}. In deep-research scenarios, difficulty is inherently linked to the number of required tool invocations—a critical dimension underexplored in current multi-hop datasets. This gap limits both performance analysis and the development of agents capable of sustained, multi-step information seeking.
    \item \textbf{Static and Constrained  Environments:} Existing evaluation frameworks predominantly rely on static, closed corpora, which fail to simulate the open-ended and unpredictable nature of the live web. The absence of dynamic environmental interaction constrains the model's capacity for autonomous exploration, strategic planning, and iterative verification—all of which are essential for navigating the complexities of deep research~\cite{gong2026strive, jia2024mill}.
\end{itemize}

To bridge these gaps, we introduce DeepResearch-9K, a challenging benchmark with different levels of difficulty designed to systematically evaluate and advance agentic capabilities in open-ended deep-research scenarios. Our work makes three core contributions:

\begin{enumerate}[leftmargin=*]
\item \textbf{Multi-level Dataset.} DeepResearch-9K provides diverse 9,000 high-quality instances across three difficulty levels ($\mathcal{L}_1$, $\mathcal{L}_2$, $\mathcal{L}_3$), with task difficulty explicitly tied to the number of required tool calls. We ensure data quality by leveraging a strong teacher model (Tongyi-DeepResearch-30B-A3B~\cite{tongyidr}) to rigorously validate solution trajectories, yielding high-quality demonstrations for supervised fine-tuning.

\item \textbf{Training Paradigm Analysis.} We systematically compare zero-RL and supervised fine-tuning followed by RL (SFT+RL) across base models (Qwen2.5-3B, Llama3.2-3B) and algorithms (PPO, GRPO). Results show supervised fine-tuning (SFT) mitigates cold-start issues, while zero-RL elicits complex reasoning, enabling smaller models to surpass larger counterparts~\cite{zeng2025simplerl, zhao2019deep}.

\item \textbf{Efficient Data Synthesis.} We propose a low-cost pipeline that extracts key entities, builds hierarchical relational graphs, and applies calibrated abstraction to generate difficulty-scaled research queries without extensive human annotation. It requires only a cost of 200 US dollars.
\end{enumerate}

\section{Overview of DeepResearch-9K} \label{sec:Overview}
\label{sec:Overview}

\begin{figure}[htbp]
    \centering
    % 定义局部颜色
    \definecolor{boxbg_yellow}{HTML}{FFF9E6}
    \definecolor{boxbg_red}{HTML}{FFE6E6}
    \definecolor{boxbg_green}{HTML}{E6F9F5}
    \definecolor{text_gray}{HTML}{333333}

    \tcbset{
        common/.style={
            enhanced,
            arc=6pt, % 稍微减小圆角以适应窄列
            boxrule=0pt,
            fontupper=\small\sffamily,
            fonttitle=\bfseries\sffamily,
            colupper=text_gray,
            width=\columnwidth, % 自动适应 SIGIR 的一栏宽度
            left=2mm,
            right=2mm,
            top=2mm,
            bottom=2mm
        }
    }

    % Question Box
    \begin{tcolorbox}[common, colback=boxbg_yellow]
        \textbf{Question:} In the sacred texts attributed to the legendary lawgiver who confronted a monarch during the Exodus, a mountain is described where divine commandments were inscribed on stone tablets. At the foot of this mountain stands an ancient monastic fortress that safeguarded one of the oldest surviving biblical manuscripts, later uncovered in the 19th century by a European scholar whose acquisition practices were widely debated. Parts of this manuscript are now preserved in a major academic library located in a city that was once divided during the Cold War. What is the name of this library?
    \end{tcolorbox}

    % \vspace{0.15cm}

        % Difficulty Box
    \begin{tcolorbox}[common, colback=boxbg_red, width=1.0\columnwidth]
        \textbf{Difficulty:} Level-3
    \end{tcolorbox}

    % \vspace{0.15cm}

    % Topic Box
% Rollout Box (Refined with Tool-use Syntax)
% Rollout Box (Refined with Tool-use Syntax)
    \begin{tcolorbox}[common, colback=boxbg_green]
        \textbf{Search Trajectory:} 
        
        \textcolor{orange}{\texttt{<Think>}}
        Lawgiver/monarch clue suggests Moses and Mt. Sinai. 
        Need to verify the fortress and manuscript at its base...
        \textcolor{orange}{\texttt{</Think>}}

        \textcolor{purple}{\texttt{<tool\_call>}} 
        search(["fortress at Mt. Sinai base", "manuscript older than empires"])
        \textcolor{purple}{\texttt{</tool\_call>}}
        
        \textcolor{blue}{\texttt{<tool\_response>}} 
        Results: St. Catherine's Monastery, Codex Sinaiticus.
        \textcolor{blue}{\texttt{</tool\_response>}}
        
        \textcolor{orange}{\texttt{<Think>}}
        The 19th-century scholar is likely Tischendorf. 
        Where are his acquisitions now? Especially in a "divided city"...
        \textcolor{orange}{\texttt{</Think>}}
        
        \textcolor{purple}{\texttt{<tool\_call>}} 
        google\_scholar(["Tischendorf fragments current locations", "divided city Cold War library"])
        \textcolor{purple}{\texttt{</tool\_call>}}
        
        \textcolor{blue}{\texttt{<tool\_response>}} 
        Leipzig University Library (Leipzig, former East Germany).
        \textcolor{blue}{\texttt{</tool\_response>}}
        
        \textcolor{orange}{\texttt{<Think>}}
        Leipzig fits the "divided" context. Conclusion reached.
        \textcolor{orange}{\texttt{</Think>}}
    \end{tcolorbox}

    % \vspace{0.15cm}

    % Final Answer Box
    \begin{tcolorbox}[common, colback=boxbg_yellow, width=1.0\columnwidth]
        \textbf{Final Answer:} Leipzig University Library
    \end{tcolorbox}
    \vspace{-3mm}
    % \vspace{0.15cm}
    \caption{A data sample from DeepResearch-9K including question, difficulty, search trajectory, and final answer.}
    \label{fig:deepresearch_sample}
    \vspace{-6mm}
\end{figure}

\subsection{Data Sample}
Figure~\ref{fig:deepresearch_sample} shows a representative data sample to provide a concrete illustration of the DeepResearch-9K dataset. Specifically, this sample exemplifies the intricate, multi-step reasoning and structured information processing required by tasks in our dataset:

\begin{itemize}[leftmargin=*]
    \item \textbf{Question}. This question involves strong entity obfuscation and requires multi-step reasoning to resolve. In DeepResearch-9K, task difficulty scales with the level of entity obfuscation and the required reasoning depth.

    \item \textbf{Difficulty}. We assume that the more difficult the problem, the more times the search tool is called. Therefore, each instance is annotated with a predefined difficulty level (from $\mathcal{L}_1$ to $\mathcal{L}_3$) measured by the number of tool calls required to resolve the task.

    \item \textbf{Search Trajectory}. We collect the search trajectory with reasoning chains from an advanced deep-research agent Tongyi-DeepResearch-30B-A3B~\cite{tongyidr}, which contains multi-round tool interactions and a step-by-step thinking process. These trajectories serve as high-quality demonstrations for supervised fine-tuning and provide rich intermediate reasoning signals for reinforcement learning. Throughout this iterative interaction, the agent utilizes a \texttt{<Think>} block to perform internal reasoning and strategic planning before invoking the \texttt{<tool\_call>} command to trigger external search tools. By alternating between these steps, the agent progressively constructs a coherent reasoning chain that bridges information gaps to reach the final answer.

    \item \textbf{Final Answer}. Each sample has a verifiable final answer. Specifically, in the following, LLM-as-Judge is adopted to verify the correctness of the answer. 
\end{itemize}

\begin{comment}
This sample illustrates how DeepResearch-9K advances beyond shallow multi-hop QA by breaking down complex problems, engaging in long-horizon tool interaction, and integrating information from multiple sources. Collectively, these features establish a rigorous testbed for deep research capabilities.
\end{comment}

\subsection{Tool-use Statistics}
\begin{figure}[t] % 去掉星号，改为单栏显示；[t] 表示放在页面顶部
\setlength\abovecaptionskip{0.2\baselineskip}
\setlength\belowcaptionskip{0.2\baselineskip}
\centering
    % 将宽度改为当前栏宽的 100% 
    \includegraphics[width=1.0\columnwidth]{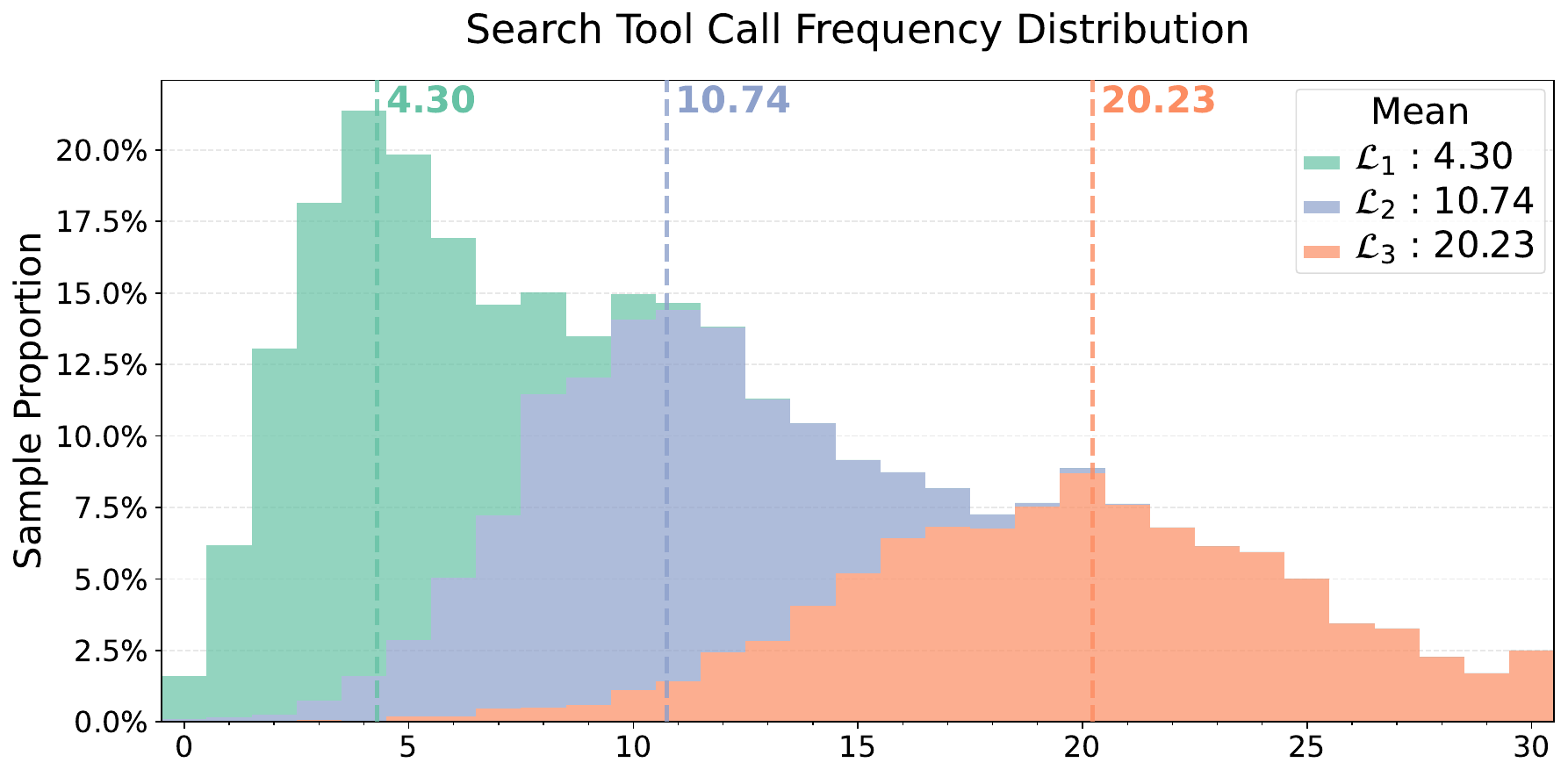}
    \caption{Distribution and mean of search tool call frequency across different difficulty levels ($\mathcal{L}_1$--$\mathcal{L}_3$) in the DeepResearch-9K dataset.}
    \label{fig:Search_count}
    \vspace{-3mm} % 如果单栏显示不需要这么大的负间距，可以注释掉或缩小
\end{figure}

To quantify the difficulty of DeepResearch-9K more precisely, we analyze the distribution of tasks across three predefined difficulty levels ($\mathcal{L}_1$, $\mathcal{L}_2$, $\mathcal{L}_3$) and their corresponding tool-interaction patterns. As illustrated in Figure~\ref{fig:Search_count}, the dataset exhibits a clear escalation in research requirements:

\textbf{Hierarchical Composition.} The dataset maintains a balanced distribution across three difficulties with 3,000 instances per level. While the logical design of these tiers is detailed in Section~\ref{sec:preliminary}, their statistical profiles already reveal a significant divergence in task nature. More precisely, $\mathcal{L}_1$ serves as a solid baseline for factual grounding, while $\mathcal{L}_2$ and $\mathcal{L}_3$ transition into multi-step relational bridging and complex research simulation with highly obfuscated entity information, respectively.

\textbf{Tool-use Intensity.} We adopt the number of autonomous searches as the primary target for research difficulty, as it directly reflects the agent's required information-seeking steps and reasoning depth, consistent with prior work~\cite{jin2025search, yang2018hotpotqa}. Our statistical analysis reveals a sharp progression in agent behavior across the hierarchy:

\begin{itemize}[leftmargin=*]
    \item Analysis of search tool call frequency:
    % \textbf{\wyh{Search Depth Analysis? Do not create the inexact words!}:} 
    The mean number of tool calls increases substantially with each level: from 4.30 at $\mathcal{L}_1$ to 10.74 at $\mathcal{L}_2$, and reaching 20.23 at $\mathcal{L}_3$. This consistent increase validates the effectiveness of our dataset construction strategy, which effectively prevents agents from bypassing reasoning steps through single-source retrieval.
    
    \item Behavioral variance: As shown in the distribution patterns (Figure~\ref{fig:Search_count}), $\mathcal{L}_1$ tasks exhibit a concentrated, low-variance distribution, indicating predictable retrieval paths. By comparison, $\mathcal{L}_2$ and $\mathcal{L}_3$ show broader distributions that tend to higher tool-use counts. This suggests that higher-level tasks not only require more steps but also introduce greater stochasticity and variability in the strategic paths agents must explore to reach the target entity.
\end{itemize}
\section{Construction of Dataset} \label{sec:preliminary}
This section presents the complete hierarchical construction strategy for the DeepResearch-9K dataset. Specifically, we design a novel multi-level difficulty scaling approach, which progressively increases task difficulty across four key dimensions: search tool call frequency, logical chain difficulty, synonymous substitution, and the degree of entity obfuscation. 

The prompt template of constructing instance of DeepResearch-9K is provided in Figure~\ref{fig:l3_prompt_template} taking $\mathcal{L}_3$ as an example due to space limit. For each level, we define separate fact-generation and QA-generation prompts, along with progressive obfuscation strategies. $\mathcal{L}_1$ focuses on entity substitution; $\mathcal{L}_2$ removes proper names, dates, and locations; $\mathcal{L}_3$ establishes a multi-layered relational chain from the seed entity to the final target, necessitating long-horizon strategic planning and step-by-step reasoning~\cite{wang2023plate, fu2025unified}. 

Besides, we introduce \textbf{DeepResearch-Hard}, a challenging subset of DeepResearch-9K with 3,974 instances. Specifically, it only contains the samples on which the teacher model Tongyi-DeepResearch-30B-A3B fails to provide a correct answer, as verified by our LLM-as-Judge framework~\cite{li2025preference,xiao2025assessment}. The complete LLM-as-Judge prompt template is provided on our code website.
Notably, these instances represent the most difficult cases in our dataset, where even an advanced deep-research agent cannot reliably produce accurate trajectories provided multi-step tool invocation~\cite{liu2025llmemb}. Thus, this subset can rigorously evaluate the limits of deep-research agents.

\begin{figure}[t]
\centering
\setlength\abovecaptionskip{0.1\baselineskip}
\begin{tcolorbox}[
  colback=blue!2!white,
  colframe=blue!50!black,
  title=Data Synthesis of DeepResearch-9K,
  coltitle=white,
  fonttitle=\bfseries,
  colbacktitle=blue!50!black,
  width=\linewidth,
  enhanced,
  breakable
]

% --- Level 3: Deep-Research ---
\noindent
\textbf{\large \textcolor{blue!70!black}{$\mathcal{L}_3$: Deep-Research (Long-Chain Reasoning)}} \
\vspace{6pt}

\small
\textbf{Fact-Generation Prompt:} \\
\texttt{Start from '\{seed\_entity\}'. Construct a 'Relay Race' chain of 5–6 entities (A$\rightarrow$B$\rightarrow$C$\rightarrow$D$\rightarrow$E$\rightarrow$Target).} \
\begin{itemize}[leftmargin=*]
    \item \textbf{Hard Link Rule:} Each hop must require a new independent search query.
    \item \textbf{Verification Constraint:} No single Wikipedia page/article may contain more than two consecutive entities.
    \item \textbf{Target Specification:} The final entity must be a specific, verifiable name.
\end{itemize}

\vspace{8pt}
\hrule
\vspace{8pt}

\textbf{QA-Generation Prompt:} \\
\texttt{Task: $\mathcal{L}_3$. Deep-Research.} \
\begin{itemize}[leftmargin=*]
    \item \textbf{Objective:} Encode the relay chain into a single, dense narrative paragraph.
    \item \textbf{Answer Format:} Output only the final target entity name.
    \item \textbf{Style Guidelines:} Use sophisticated, nested sentence structures (e.g., relative clauses with ``whose'', ``where'', ``that'').
    \item \textbf{Obfuscation Strategy:} Apply high-level obfuscation throughout:
    \begin{enumerate}[leftmargin=*]
        \item \textbf{No proper names:} Replace with functional roles (e.g., ``the telecommunications expert'').
        \item \textbf{No exact dates:} Use historical eras (e.g., ``throughout the industrial expansion'').
        \item \textbf{No exact locations:} Use geographical/cultural descriptions (e.g., ``the central heartland'').
    \end{enumerate}
    \item \textbf{Difficulty Target:} The narrative should necessitate $\geq 15$ independent search queries to resolve.
    \item \textbf{Forbidden Elements:} No lists, no sequential markers, no arrows ($\rightarrow$).
    \item \textbf{Quality Exemplar:} Interweave historical conflict, cultural heritage, and technical expertise into a single coherent puzzle.
\end{itemize}

\end{tcolorbox}
\caption{Detailed prompt template for Level-3 ($\mathcal{L}_3$) deep-research tasks in the DeepResearch-9K dataset.}
\label{fig:l3_prompt_template}
\vspace{-3mm}
\end{figure}

The construction process comprises four primary stages: 1) seed entity identification and extraction, 2) progressive reasoning chain construction, 3) progressive entity obfuscation, and 4) rule-based quality assurance. These steps are detailed in the following sections.
Overall, our pipeline delivers DeepResearch-9K: a high-quality, multi-round interaction resource crafted for deep-research to advance agentic reasoning. The entire procedure requires 200 USD in Serper API fees and approximately 8,064 A100 GPU hours.

\subsection{Seed Entity Identification and Extraction}
Our selection of three complementary benchmarks—HotpotQA~\cite{yang2018hotpotqa}, 2WikiMultihopQA~\cite{ho2020constructing}, and MuSiQue~\cite{trivedi2022musique} is designed to address distinct aspects of complex question answering. Specifically:

% \begin{itemize}[leftmargin=*]
%     \item \textbf{2WikiMultihopQA} provides structurally explicit reasoning chains with Wikipedia-grounded verifiability~\cite{chen2020open}, offering reliable factual anchors for our seed entity extraction. Its multi-hop annotations ensure logical consistency in constructed chains~\cite{qi2019answering}.
    
%     \item \textbf{HotpotQA} serves as a quality benchmark with its human-verified question-answer pairs and diverse reasoning types (e.g., bridging, comparison)~\cite{min2019multi}. We leverage its question diversity to enrich our entity extraction pool. 
    
%     \item \textbf{MuSiQue} contributes rigorous anti-cheating design and real challenging multi-hop instances~\cite{zhou2021towards}. Its construction methodology, which prevents shortcut solutions~\cite{geva2021did}, directly guides our rule-based validation implementation and obfuscation strategies.
% \end{itemize}

\begin{itemize}[leftmargin=*]
    \item \textbf{2WikiMultihopQA} contributes structurally explicit reasoning chains with clear entity boundaries~\cite{qi2019answering}.
    
    \item \textbf{HotpotQA} provides human-verified QA pairs with diverse reasoning types and multi-hop annotations~\cite{min2019multi}.
    
    \item \textbf{MuSiQue} offers rigorous anti-cheating design and challenging multi-hop instances~\cite{zhou2021towards}.
\end{itemize}

% These foundations ensure that DeepResearch-9K effectively inherits: (1) structural clarity from 2WikiMultihopQA, (2) quality assurance from HotpotQA, and (3) rigorous validation from MuSiQue, while addressing their individual limitations through our hierarchical difficulty scaling. 
These foundations ensure DeepResearch-9K inherits structural clarity, quality assurance, and rigorous validation. From each dataset, we randomly sample 1,000 instances as the initial synthesis pool. Using DeepSeek-V3~\cite{liu2025deepseek}, we extract key entities (e.g., names, locations, time nodes) from each sample as initial nodes for a relational network. By iteratively broadening entity associations, we construct a multi-hop reasoning graph~\cite{xiong2020answering}. Selective path extraction from this graph enables explicit control over reasoning horizon and task difficulty~\cite{min2019multi,shen2020modeling}, allowing difficulty escalation that stimulates agents to navigate increasingly intricate reasoning chains.

\subsection{Progressive Reasoning Chain Construction}
\label{sec:Progressive}
To effectively differentiate task difficulty, we impose varying constraints on the reasoning length and the required search frequency for each level ($\mathcal{L}_1$ to $\mathcal{L}_3$), ensuring a measurable progression in agent performance requirements:

\begin{itemize}[leftmargin=*]
    \item \textbf{$\mathcal{L}_1$ : } \textbf{Direct Attribute Mapping}. For $\mathcal{L}_1$ tasks, the construction begins by extracting a seed entity and identifying three specific factual attributes~\cite{ling2015design}. To introduce initial difficulty, we mandate the inclusion of at least one lesser-known synonym or technical term to describe these attributes, thereby testing the agent's basic semantic grounding. The corresponding QA pairs are strictly constrained to 1--2 search iterations.
    
    \item \textbf{$\mathcal{L}_2$ : } \textbf{Multi-hop Relational Bridging}. At this level, we escalate task difficulty by constructing an explicit multi-hop $A \rightarrow B \rightarrow C$ relational chain. Unlike the direct attribute mapping in $\mathcal{L}_1$, $\mathcal{L}_2$ tasks introduce deliberate informational gaps between the initial query and final target~\cite{welbl2018constructing}. The agent must identify and link multiple intermediate entities, thereby preventing resolution through a single retrieval step. Consequently, $\mathcal{L}_2$ questions require extended search trajectories to resolve intermediate nodes and reach the target entity. 
    
    \item \textbf{$\mathcal{L}_3$ : } \textbf{Simulating Deep Research}. At its most difficult, the benchmark simulates authentic research via an extended sequential chain ($A \rightarrow B \rightarrow \dots \rightarrow \text{Target}$)~\cite{bai2021multi}. The core of this level is a strict verification constraint: each transitional jump between entities must require a new, independent search query~\cite{chen2019evaluating, lamm2021qed}. To ensure this, our construction pipeline verifies that no single knowledge source (e.g., a Wikipedia page) contains more than two consecutive entities from the chain. Furthermore, the task is presented as a dense narrative inquiry using sophisticated, nested sentence structures (e.g., relative clauses) to integrate these dependencies into a single cohesive paragraph. This design effectively removes shortcut reasoning paths, ensuring a minimum of 15 search iterations for successful resolution~\cite{mishra2022lila}. This level specifically challenges the agent’s capacity for sustained strategic planning and autonomous information synthesis across fragmented web sources.
\end{itemize}

\subsection{Progressive Entity Obfuscation}
\label{sec:Obfuscation}
To simulate the ambiguity and information sparsity characteristic of real-world research, we employ a Progressive Entity Obfuscation strategy. This approach systematically masks explicit identifiers, compelling the agent to depend on semantic reasoning and cross-source verification instead of straightforward keyword matching.

\begin{itemize}[leftmargin=*]
    \item \textbf{Light Obfuscation ($\mathcal{L}_1$) :} We apply fundamental transformations by substituting primary entities with closely related descriptions or synonyms (e.g., replacing "Beijing" with "the capital of China")~\cite{liu2020logiqa}.
    
    \item \textbf{Moderate Obfuscation ($\mathcal{L}_2$) :} We escalate difficulty by prohibiting proper names, exact dates, and specific locations for 1–2 key entities in the reasoning chain. Entities are replaced with descriptive roles (e.g., "the mathematician who won the first Fields Medal") or general spatio‑temporal markers (e.g., "a European capital city" in the "early 20th century")~\cite{yao2021refining}.
    
    \item \textbf{High Obfuscation ($\mathcal{L}_3$) :} At the highest level, we employ comprehensive masking across the entire inquiry. All proper identifiers are stripped and replaced with nested descriptions and historical eras (e.g., "the mid-Victorian age"). These entities are embedded in a dense narrative using complex syntactic structures such as relative clauses~\cite{wu2018learning}. This obfuscation ensures that no single identifier can be resolved without multiple iterative search queries.
\end{itemize}

\subsection{Rule-based Quality Assurance}
To ensure data integrity and format consistency, we implement a rule-based validation pipeline that enforces the construction constraints defined in Sections~\ref{sec:Progressive} and Sections~\ref{sec:Obfuscation}~\cite{gehrmann2021gem}. We establish quality standards by deconstructing high-quality examples from BrowseComp-Plus~\cite{chen2025browsecomp} using LLM-based analysis, from which we derive explicit construction guidelines~\cite{kreutzer2022quality, thakur2021beir, wu2025empowering}, ensuring logical rigor and consistent presentation quality.

%\vspace{\baselineskip}
This multi-stage construction pipeline, coupled with rigorous cross-document independence checks and automated format validation, ensures that every task in DeepResearch-9K possesses a factual grounding that is robustly verifiable while maintaining an extreme level of research difficulty for autonomous agents.
\section{Experiments} \label{sec:Experiments}
\subsection{Experimental Setting}
\subsubsection{Datasets} \label{sec:exp:dataset}
In the following experiments, on the one hand, the deep-research ability of the agent is evaluated on all samples of DeepResearch-9K dataset.
Besides, we additionally evaluate on BrowseComp-Plus~\cite{chen2025browsecomp}, a famous deep-research benchmark dataset which comprises 830 complex tasks requiring strategic information seeking and comparative analysis.

On the other hand, for training paradigm analysis, we combine all 5,026 correct trajectories with a randomly selected subset of 2,200 incorrect samples (from DeepResearch-Hard), forming a training set of 7,226 instances. The remaining 1,774 samples from DeepResearch-Hard form the test set.

\begin{comment}
We construct our dataset by randomly sampling 1,000 question-answer pairs from each of HotpotQA~\cite{yang2018hotpotqa}, 2WikiMultihopQA~\cite{ho2020constructing}, and MuSiQue~\cite{trivedi2022musique}. For each seed pair, we synthesize tasks across three levels of increasing complexity (L1–L3), yielding a total of 9,000 multi-level research questions. Using LLM-as-a-judge with DeepSeek-V3~\cite{liu2025deepseek}, we evaluate teacher-model trajectories against ground truth, identifying 5,026 correct samples. The remaining 3,974 incorrect samples are designated as \textbf{DeepResearch-Hard} and reserved for further evaluation. 

\end{comment}
    
\subsubsection{Models and Training} 
Following the construction of 9,000 synthetic samples, we employ the Tongyi-DeepResearch-30B-A3B model~\cite{tongyidr} as a teacher model to answer each question and generate complete detailed search trajectories, which include search queries, intermediate reasoning steps, and final predictions. To create a structured knowledge base from these trajectories, we utilize the DeepSeek-V3 API~\cite{liu2025deepseek} to generate Wikipedia-style summaries, forming a searchable corpus that is indexed to facilitate efficient retrieval during subsequent training phases. Our training paradigm is a two-stage process centered on Reinforcement Learning (RL)~\cite{kaelbling1996reinforcement}, built upon the Search-R1 framework~\cite{jin2025search}:

\begin{itemize}[leftmargin=*]
    \item \textbf{Zero-RL}: In the first stage, we directly apply RL to the base (non-instruction-tuned) language model using the entire training set. This stage aims to elicit and bootstrap foundational reasoning and tool-use capabilities from the beginning.
    \item \textbf{SFT + RL}: The second stage begins with Supervised Fine-Tuning (SFT) on the subset of correct trajectories to explicitly learn the teacher model's reasoning patterns and tool-calling strategies~\cite{chittepu2025ml}. After being initialized via supervised fine-tuning, the model then takes a second round of RL training on the entire training set to further refine and strengthen the acquired behaviors~\cite{ouyang2022traininglanguagemodelsfollow}.
\end{itemize}

Both stages are implemented using a combination of Group Relative Policy Optimization (GRPO)~\cite{guo2025deepseek,shao2024deepseekmath} and Proximal Policy Optimization (PPO)~\cite{schulman2017proximal} algorithms. The reward signal in both RL phases is provided by DeepSeek-V3~\cite{liu2025deepseek}, which evaluates the quality and correctness of the agent's reasoning trajectories. This hybrid strategy leverages GRPO's advantage in multi-objective, group-based reward scenarios alongside PPO's reliable policy updates. We apply both zero-RL and SFT+RL on two base model architectures: Qwen-2.5-3B~\cite{qwen2} and Llama-3.2-3B~\cite{dubey2024llama}.

\subsubsection{Evaluation Metrics}

To comprehensively and rigorously evaluate the performance of our deep-research agent, we employ the following two primary metrics:

\begin{table}[t]
\centering
\setlength\abovecaptionskip{0\baselineskip}
\caption{Accuracy of Tongyi-DeepResearch-30B-A3B on DeepResearch-9K}
\label{tab:results_difficulty}
\begin{tabular}{cccccccc}
\toprule
\textbf{Level} & \textbf{Correct} & \textbf{Incorrect} & \textbf{Accuracy} \\ 
\midrule
$\mathcal{L}_1$ & 2,174 & 826  & 72.47\% \\
$\mathcal{L}_2$ & 2,140 & 860  & 71.33\% \\
$\mathcal{L}_3$ & 712   & 2,288 & 23.73\% \\ 
All & 5,026 & 3,974 & 55.84\% \\
\bottomrule
\end{tabular}
\vspace{-3 mm}
\end{table}

\begin{itemize}[leftmargin=*]
    \item \textbf{Search Count}: We calculate the cumulative number of search tool invocations (e.g., Serper API~\cite{li2025loki}) per reasoning trajectory. This metric reflects the agent's information-seeking ability and planning effectiveness in open-ended research tasks. Comparing search counts across DeepResearch-9K's difficulty tiers ($\mathcal{L}_1$–$\mathcal{L}_3$) and BrowseComp-Plus~\cite{chen2025browsecomp} helps assess whether $\mathcal{L}_1$–$\mathcal{L}_3$ exhibit clear difficulty gradation and whether $\mathcal{L}_3$ reaches a challenge level comparable to BrowseComp-Plus.

    \item \textbf{Accuracy}: We adopt an LLM-as-Judge paradigm~\cite{gu2024survey} using DeepSeek-V3~\cite{liu2024deepseek} to evaluate final answer correctness. We analyze accuracy across difficulty levels $\mathcal{L}_1$–$\mathcal{L}_3$ on DeepResearch-9K, as well as overall accuracy on BrowseComp-Plus (830 tasks). This evaluation assesses model robustness against progressive entity obfuscation and complex relational reasoning.
\end{itemize}

\subsection{Direct Evaluation}
% However, the accuracy of Tongyi-DeepResearch-30B-A3B agent on our DeepResearch-Hard dataset is 0. Consequently, 

We first evaluate the Tongyi-DeepResearch-30B-A3B~\cite{tongyidr} model on our proposed benchmark dataset. As shown in Table~\ref{tab:results_difficulty}, it achieves 72.47\% accuracy on $\mathcal{L}_1$, demonstrating competent performance on direct factual retrieval. Accuracy slightly drops to 71.33\% on $\mathcal{L}_2$, indicating that multi-hop reasoning within two hops remains manageable for the teacher model. However, performance sharply declines to 23.73\% on $\mathcal{L}_3$, highlighting the substantial challenge posed by deep-research tasks that require multi-step reasoning chains and progressive entity obfuscation. Overall, the teacher model achieves 55.84\% accuracy across all 9,000 instances.

To further validate that the questions in $\mathcal{L}_3$ are challenging, we evaluate Tongyi-DeepResearch-30B-A3B on BrowseComp-Plus~\cite{chen2025browsecomp}, the most challenging benchmark dataset for complex web research introduced in Section~\ref{sec:exp:dataset}. The result shows that Tongyi-DeepResearch-30B-A3B achieves an accuracy of 24.94\%  on BrowseComp-Plus, closely matching its 23.73\% accuracy on $\mathcal{L}_3$ of DeepResearch-9K. This result demonstrates that $\mathcal{L}_3$ accurately captures the intricate relationships and high reasoning demands of real-world research, while also highlighting the substantial space for improvement in current deep-research capabilities.

\begin{figure} % 保持星号以支持跨栏显示
\setlength\abovecaptionskip{0.1\baselineskip}
\setlength\belowcaptionskip{0.2\baselineskip}
% \caption{Model Performance Comparison on DeepResearch-9K.}
\centering
    
    % 将宽度从 1.0\columnwidth 改为 1.0\linewidth，这样会占据整个页面的宽度
    \includegraphics[width=1.0\linewidth]{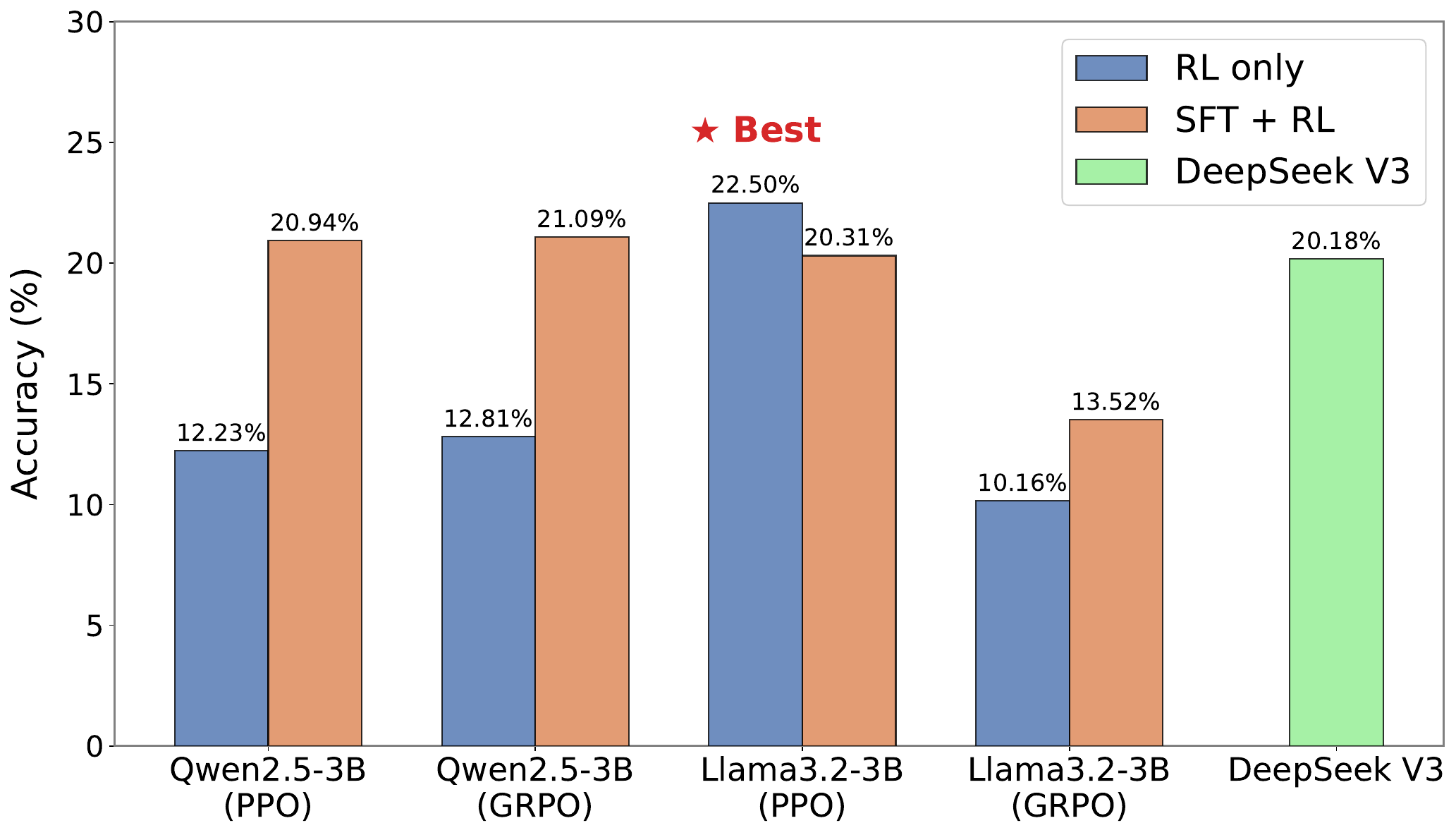}
    
    \caption{Model performance comparison on the test set of DeepResearch-9K.}
    \label{fig:model_performance}
    \vspace*{-5mm} % 图片拉大后，通常需要适当调小负间距
\end{figure}

\subsection{Training Paradigm Analysis}
To systematically investigate how different training strategies shape agent capabilities, we train Qwen-2.5-3B~\cite{qwen2} and Llama-3.2-3B~\cite{dubey2024llama})  two reinforcement learning algorithms (PPO~\cite{schulman2017proximal} and GRPO~\cite{shao2024deepseekmath}), and two training paradigms (Zero-RL and SFT+RL). 
All models are trained within our DeepResearch-R1 framework and deployed as autonomous agents, with vLLM enabling efficient inference rollouts under identical environmental conditions~\cite{kwon2023efficient, zhang2025process}.
%in Table~\ref{tab:rl_hyperparams} in Appendix~\ref{app:Experimental_Setup}.  

%\subsection{Findings}
%Building on the experimental results presented in Figure~\ref{fig:model_performance}, our analysis 
The results are shown in Figure~\ref{fig:model_performance}, contrasting Zero-RL (blue bars) with SFT+RL (orange bars). We also evaluate the DeepSeek V3 (green bar) as a competitive state-of-the-art model on the full test set. The results reveal three critical insights regarding model performance and training dynamics:

\begin{itemize}[leftmargin=*]
    \item \textbf{SFT Dependency in Qwen Models:} The Qwen-2.5-3B model exhibits a strong dependency on supervised pre-training. The "Cold Start" (Zero-RL) approach yields modest accuracy ($\sim$12\%), whereas the "Warm Start" (SFT+RL) strategy significantly boosts performance to over 20\% across both PPO and GRPO algorithms. This indicates that for this architecture, SFT is essential for grounding the agent's prior knowledge before it can effectively leverage search tools.

    \item \textbf{Peak Performance of Specialized Agents:} Remarkably, the Llama-3.2-3B model trained with PPO (Zero-RL) achieves the highest overall accuracy of 22.50\% (marked with \ding{72} in Figure~\ref{fig:model_performance}), clearly outperforming all other models, including the larger DeepSeek V3 baseline (20.18\%). This suggests that smaller, highly specialized agents can surpass closed-source reference models when their reward signals are effectively aligned through targeted reinforcement learning.
    
    \item \textbf{Benchmark Difficulty:} Despite these optimizations, model performance consistently plateaus near 20\%. DeepSeek V3 achieves only 20.18\%, clearly underscoring the substantial challenge of DeepResearch-9K and confirming that the dataset probes the upper limits of current agentic capabilities in multi-hop reasoning and information synthesis.
\end{itemize}

\section{Related work} \label{sec:Related work}
Deep-research demands the automation of complex workflows, requiring agents to actively discover and validate knowledge across sources, and iteratively synthesize insights through multi-step, open-ended exploration~\cite{zhang2025deep}. 

\subsection{Deep-Research Dataset}
Recent benchmarks have shifted from closed-book QA to interactive, web-grounded evaluation~\cite{petroni2021kilt}. GAIA~\cite{mialon2023gaia} distinguished tasks by the number of required reasoning steps and tool calls, simulating real-world question answering that often necessitates web search and multi-modal processing. GPQA~\cite{rein2024gpqa} and Humanity's Last Exam~\cite{phan2025humanity} focus on expert-level, domain-specific questions that resist simple retrieval, pushing agents toward deeper research behaviors. At the intersection of browsing and competition, BrowseComp~\cite{wei2025browsecomp} introduces a benchmark where agents must answer questions by actively navigating and comparing information across multiple websites, explicitly testing their ability to conduct comparative research under time constraints. 

Despite these advances in evaluation, a significant gap remains in training resources~\cite{liu2023agentbench}. Current benchmarks are designed primarily for assessment, not for providing the large-scale, high-quality interaction trajectories needed to train agents from scratch or iteratively improve them~\cite{chang2024agentboard}. The scarcity of such data hinders the development of agents capable of sustaining the long-horizon, strategically complex workflows that define deep-research~\cite{chen2025browsecomp,liu2023bolaa}. Our approach tackles this dual challenge by offering a scalable pipeline that synthesizes verified, multi-step training trajectories, thereby facilitating the agent training.

% Early deep research established the foundation for multi-hop reasoning within controlled, closed-domain environments. By designing intricate logical dependencies, benchmarks such as NQ~\cite{kwiatkowski2019natural}, HotpotQA~\cite{yang2018hotpotqa}, 2WikiMultihopQA~\cite{ho2020constructing}, and MuSiQue~\cite{trivedi2022musique} have demonstrably advanced the comprehension and reasoning capabilities of models. However, these tasks primarily focus on extracting answers from a predefined text corpus, operating within a static and bounded evaluation setup~\cite{petroni2021kilt}. Crucially, there remains a severe lack of large-scale, challenging training datasets that authentically capture the dynamic, multi-faceted complexity of real-world deep research tasks.~\cite{ji2025deepambigqa,ahuja2023mega}. This limitation leaves a gap for high-density supervision data that can stimulate the strategic planning and long-horizon planning required for deep research scenarios~\cite{thakur2021beir}.

\subsection{Deep-Research Framework}
% This data scarcity is compounded by a second, equally critical bottleneck: the absence of accessible, open-source frameworks for synthesizing such data and iteratively training agents. The high cost of manual annotation ~\cite{koch2022real} and the technical complexity of building robust data pipelines ~\cite{hovy2022toward} create prohibitive barriers to entry. Moreover, without systematic tools to automatically diagnose failures and generate targeted adversarial examples ~\cite{guo2024llm}, the iterative optimization of agents remains inefficient and resource-intensive. The lack of a low-cost, closed-loop “train-evaluate-adapt” pipeline ~\cite{ye2024agent} centralizes advanced research within resource-rich groups and stifles broader innovation.
The development of deep-research agents relies not only on challenging tasks but also on the frameworks that enable their training and evaluation. Recent efforts have begun to provide standardized evaluation environments. 
Frameworks like DR-Arena~\cite{gao2026dr} and DeepResearch Bench~\cite{du2025deepresearch} further formalize the evaluation of autonomous research agents by providing standardized environments for measuring performance on complex, open-ended tasks. DeepWideSearch~\cite{lan2025deepwidesearch} complements this by demanding strategic planning over long-horizon, exploratory web searches, evaluating both depth and breadth of investigation. On the training aspect, projects like ToolBench~\cite{qin2023toolllm} show the value of tool-use trajectories, while more recent work such as Search-R1~\cite{jin2025search} synthesizes “thought-action” trajectories to distill robust policies. This situation emphasizes a broader absence of accessible, open-source frameworks dedicated to synthesizing training data and iteratively improving agents~\cite{ji2025deepambigqa,ahuja2023mega}. In contrast, our framework is explicitly designed as a reusable, open-source pipeline that automates the generation of challenging data.
\section{Conclusion}
Faced with the gap between constrained QA and open-ended research, we introduce DeepResearch-9K, a challenging benchmark dataset for evaluating agentic deep-research capabilities in this work. 
Specifically, we release the whole pipeline of dataset construction without extensive human annotation, improving accessibility and scalability.
Besides, we develop an open-source training framework DeepResearch-R1 for advanced deep-research agent training.
%It features difficulty scaling tied to tool-call requirements, progressive entity obfuscation for realistic complexity, and teacher-verified trajectories for process-level supervision. Our efficient construction pipeline generates challenging queries without extensive human annotation, improving accessibility and scalability. 
Our experiments show that: 1) direct evaluation performance of Tongyi-DeepResearch-30B-A3B justifies our benchmark dataset poses considerable difficulty and significant challenges and 2) the model trained on our dataset achieves state-of-the-art deep-research performance against DeepSeek V3. 
%The performance plateau across models confirms the benchmark's challenge and its effectiveness in probing current reasoning limits. 
%We release DeepResearch-9K to bridge the gap between constrained QA and open-ended research, 
To sum up, this work provides both an evaluation benchmark and a training framework to advance a more powerful deep-research agent.

\section*{ACKNOWLEDGEMENT}

This research was partially supported by National Natural Science Foundation of China (No.62502404), Hong Kong Research Grants Council (Research Impact Fund No.R1015-23, Collaborative Research Fund No.C1043-24GF, General Research Fund No. 11218325), Institute of Digital Medicine of City University of Hong Kong (No.9229503), Huawei (Huawei Innovation Research Program), Tencent (Tencent Rhino-Bird Focused Research Program, Tencent University Cooperation Project), Didi (CCF-Didi Gaia Scholars Research Fund), Kuaishou (CCF-Kuaishou Large Model Explorer Fund No. 2025008, Kuaishou University Cooperation Project), Bytedance, Shenzhen Technology University Discipline Enhancement Project (2025–2027) under Grant No. 2024ZDJS063, and Shenzhen Technology University School-level Research Project (2025–2027) under Grant No. 20251061020002.

\bibliographystyle{ACM-Reference-Format}
\bibliography{7-reference}

%\appendix
%\input{8.appendix}

\end{document}